\documentclass[letterpaper]{article} 
\usepackage{aaai24}  
\usepackage{times}  
\usepackage{helvet}  
\usepackage{courier}  
\usepackage[hyphens]{url}  

\usepackage{graphicx} 
\usepackage{natbib}  
\usepackage{caption} 
\usepackage{algorithm}
\usepackage{algorithmic}
\usepackage{amsmath}
\usepackage{amssymb}
\usepackage{booktabs}
\usepackage{array}
\usepackage{newfloat}
\usepackage{listings}
\DeclareCaptionStyle{ruled}{labelfont=normalfont,labelsep=colon,strut=off} 
\floatstyle{ruled}
\newfloat{listing}{tb}{lst}{}
\floatname{listing}{Listing}
\title{Self-supervised Learning for Enhancing Geometrical Modeling \\ in 3D-Aware
Generative Adversarial Network}

\author{
Jiarong Guo$^{1}$ \quad Xiaogang Xu$^{2,3}$ \quad Hengshuang Zhao$^{4}$ \\
$^1$ HKUST \quad $^2$ Zhejiang Lab \quad $^3$ Zhejiang University \quad $^4$ HKU\\
{\tt \small jguoaz@connect.ust.hk, xgxu@zhejianglab.com, hszhao@cs.hku.hk}
}

\usepackage{bibentry}

\begin{document}

\maketitle

\begin{abstract}
3D-aware Generative Adversarial Networks (3D-GANs) currently exhibit artifacts in their 3D geometrical modeling, such as mesh imperfections and holes. These shortcomings are primarily attributed to the limited availability of annotated 3D data, leading to a constrained "valid latent area" for satisfactory modeling. To address this, we present a Self-Supervised Learning (SSL) technique tailored as an auxiliary loss for any 3D-GAN, designed to improve its 3D geometrical modeling capabilities. Our approach pioneers an inversion technique for 3D-GANs, integrating an encoder that performs adaptive spatially-varying range operations. Utilizing this inversion, we introduce the Cyclic Generative Constraint (CGC), aiming to densify the valid latent space. The CGC operates via augmented local latent vectors that maintain the same geometric form, and it imposes constraints on the cycle path outputs, specifically the generator-encoder-generator sequence. This SSL methodology seamlessly integrates with the inherent GAN loss, ensuring the integrity of pre-existing 3D-GAN architectures without necessitating alterations. We validate our approach with comprehensive experiments across various datasets and architectures, underscoring its efficacy. Our project website: \url{https://3dgan-ssl.github.io}.
\end{abstract}
\section{Introduction}

{\bf Background.} 3D Generative Adversarial Networks (3D-GANs) have recently gained prominence, finding applications across various fields, notably in generating multi-view human and animal faces~\cite{or2022stylesdf, EG3D, GET3D}. Unlike their 2D counterparts~\cite{SuperRes-1, SuperRes-2, SuperRes-Survey}, 3D-GANs derive 3D feature representations from Gaussian noise and leverage implicit Neural Radiance Fields (NeRF)\cite{nerf} to render new perspectives. Additionally, these networks can produce 3D meshes via density modeling\cite{or2022stylesdf}.

\begin{figure}
    \centering
    \centerline{\includegraphics[width=1\linewidth]{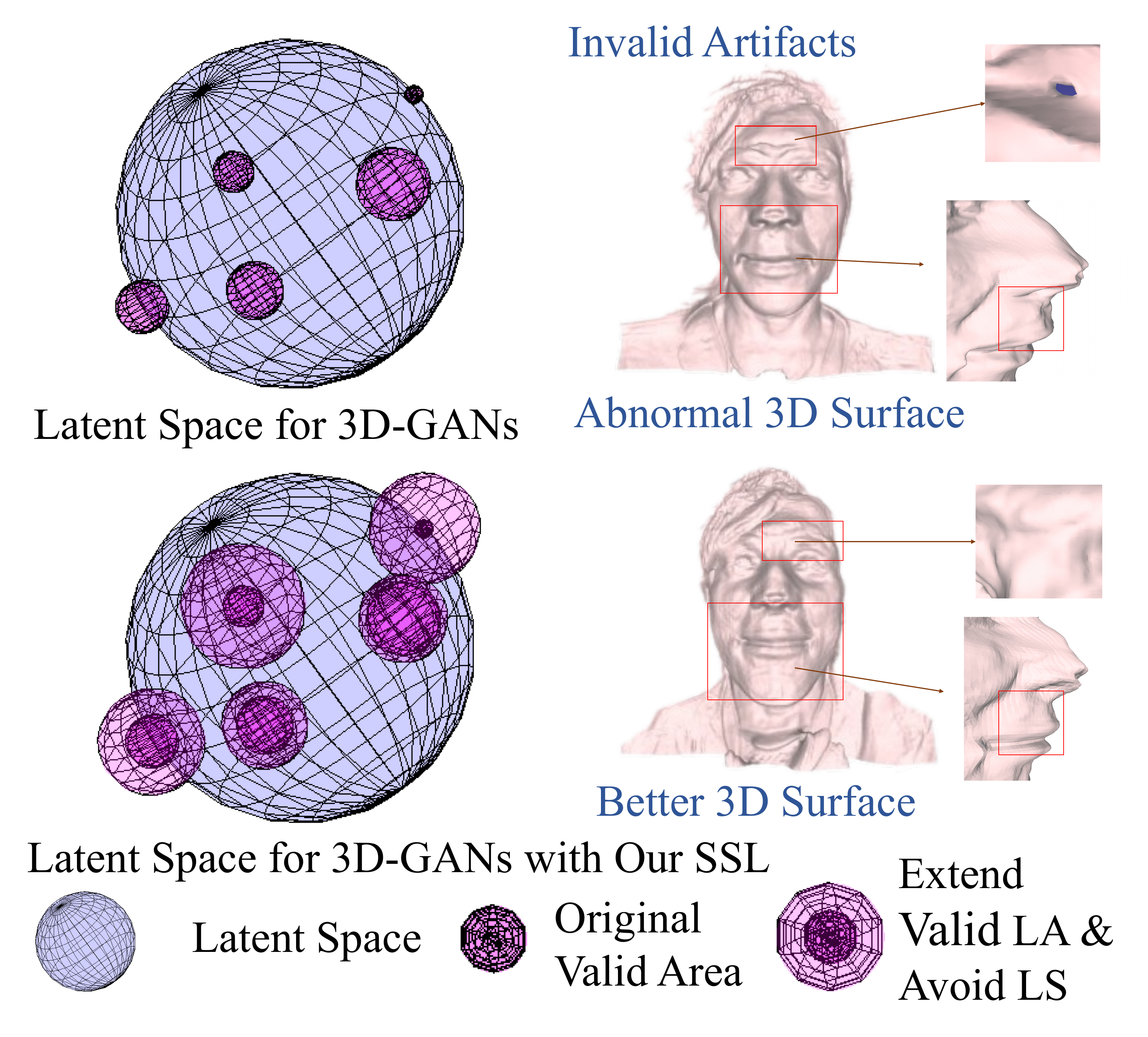}}
    \vspace{-0.2in}
    \caption{\small The motivation of our work. ``LA" means Latent Areas, and ``LS" denotes Local Saltation in the 3D representation. The latent spaces of existing 3D-GANs have discrete valid latent areas that can lead to admired geometrical modeling, due to the lack of 3D supervision.
    Our proposed CGC can extend the valid latent areas. Take EG3D~\cite{EG3D} on FFHQ~\cite{kazemi2014one} as an example here.
    }
    \label{fig:teaser}
    \vspace{-0.25in}
\end{figure}

{\bf Challenge.} Predominantly, 3D-GANs utilize 2D-based losses, focusing on rendered image quality. While existing 3D regularization losses address concerns like the visibility and validity of the Signed Distance Function (SDF)\cite{or2022stylesdf, GET3D}, they fall short in directly supervising geometrical modeling. This limitation arises from the sparse 3D annotations in prevalent datasets like FFHQ\cite{style-FFHQ} and AFHQ~\cite{stargan-AFHQ}. Consequently, current 3D-GANs often yield flawed mesh outputs, exhibiting holes, missing components, and irregularities (see Fig.~\ref{fig:teaser}).

{\bf Observation.} In this paper, we introduce a Self-Supervised Learning (SSL) framework aimed at enhancing the geometrical modeling of general 3D-GANs. We observe that latent spaces in 3D-GANs trained solely with 2D supervision tend to encompass \textit{discrete} ``valid latent areas" (i.e., the latent space which can lead to satisfactory geometrical modeling via the generator). 
A valid latent point $z$ might be encompassed by invalid neighboring points, as depicted in Fig.~\ref{fig:teaser} (this observation is visualized by adding local Gaussian perturbations to latent points, and then analyzing the outputs). 
This phenomenon arises from the limitations of 2D supervision, which cannot effectively supervise all viewpoints of a 3D model.

{\bf Motivation.} To address this challenge, we propose a solution that reduces abrupt shifts in the geometrical representation of local points within the latent space. This alteration enlarges the valid latent areas, consequently creating a dense latent space where all latent points $z$ yield desired 3D modeling outcomes. Our approach, known as Cyclic Generative Constraint (\textbf{CGC}), is designed to achieve this expansion.

{\bf Technical novelty: 3D inversion and CGC.} 
With a given latent $z$, the generator $\mathcal{G}$ produces a 3D feature representation $r$, determining the shape of the resulting 3D mesh. When combined with a camera pose $p$, $r$ enables generating a rendered image $I_p$.
To mitigate abrupt shifts in the geometrical representation of local points, our approach first involves obtaining the local neighbor $z^*$ of $z$ with identical geometrical representation. Subsequently, we enforce the 3D outputs of both $z$ and $z^*$ to match. 

To ensure this consistent geometrical representation, we employ an inversion strategy rather than solely introducing local perturbations to derive $z^*$.
Diverging from established 3D-GAN inversion techniques that rely on the rendered image $I_p$ as input for inversion, our approach introduces an inversion encoder $\mathcal{E}$ where the input is the 3D representation $r$, resulting in $z^*=\mathcal{E}(r)$. Conventional 2D inversion methods encounter a challenge in accurately formulating local samples due to the deficiency of comprehensive 3D information. In contrast, our 3D inversion approach circumvents this limitation, ensuring geometrical coherence between $z$ and $z^*$.
Optimizing $\Vert z - z^* \Vert$ guarantees the proper convergence of the inversion encoder. Moreover, to ensure high-quality inversion, we stress the significance of incorporating spatial-varying operations within the encoder.

Another 3D representation $r^*$ can be obtained as $r^*=\mathcal{G}(z^*)$.
Minimizing $\Vert r - r^* \Vert$ mitigates sudden shifts in the synthesized 3D representation at the latent point $z$, which is the implementation of CGC. $\mathcal{G}$ and $\mathcal{E}$ can be alternately optimized, leading to dual convergence. This process culminates in achieving great 3D synthesis across the entire latent space, resulting in a densely valid latent space.

{\bf Contribution.} The designed SSL approach can seamlessly integrate as a plug-and-play module into existing 3D-GANs.
Our contributions are as listed as follows:
\begin{itemize}

\item We propose a novel SSL strategy to enhance the geometrical modeling in 3D-GANs, by using the cycle constraint between the 3D feature space and the latent space to weaken the synthesis of abnormal 3D models.

\item
We design a new 3D-GAN inversion scheme to complete the optimization of CGC, guaranteeing the integrality of geometrical information during the inversion and generation via spatial-varying operations.

\item Extensive experiments are conducted on public datasets, demonstrating the effectiveness of our proposed SSL methods in improving 3D modeling qualitatively and quantitatively. Significantly, the corresponding 2D synthesis performance is also improved with the enhancement of 3D modeling.

\end{itemize}

\begin{figure*}[t]
    \centering
    \centerline{\includegraphics[width=1\linewidth]{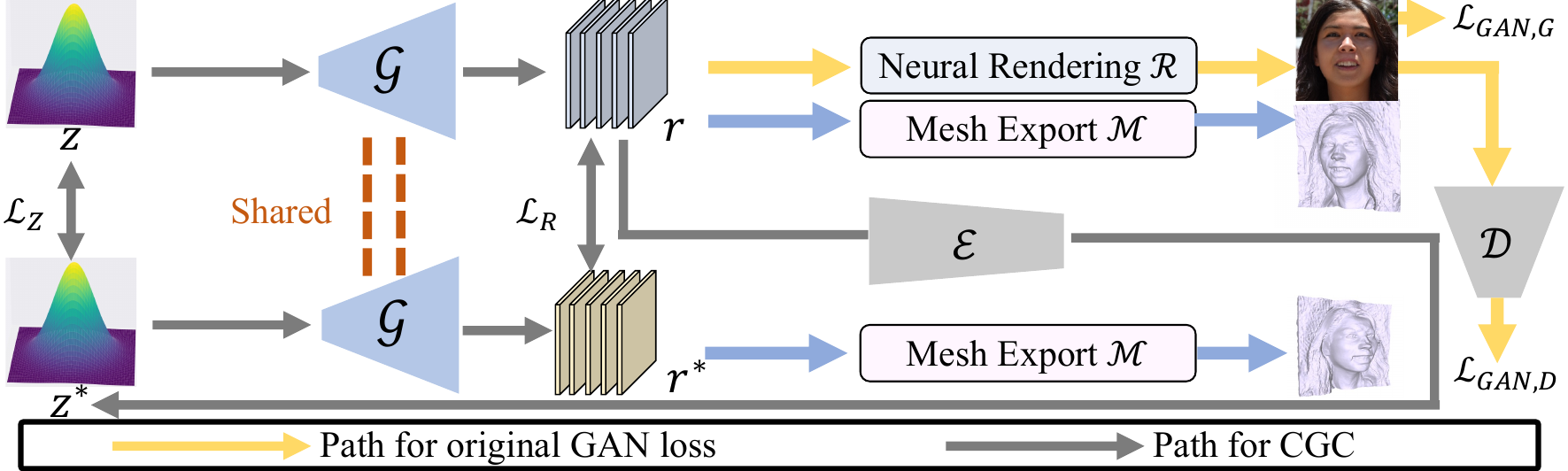}}
    \vspace{-0.1in}
    \caption{
    \small Our framework's architecture involves the cycle path (shown in gray) of the generator-encoder-generator, which embodies the CGC. CGC enforces valid latent regions conducive to effective 3D modeling, by constraining the 3D representation of $z$ and $z^*$ that have the same geometrical feature to closely match. The encoder's structure is illustrated in Fig.~\ref{fig:framework-encoder}.
    }
    \label{fig:framework}
    \vspace{-0.2in}
\end{figure*}

\section{Related Works}

\noindent\textbf{3D Geometry Learning.}
The field of 3D modeling has extensively explored various representations, such as voxels, meshes, and point clouds, leading to a myriad of applications~\cite{3dsurvey, pcsurvey, point-survey}. A notable advancement has been the success of Neural Radiance Fields (NeRF), which sparked interest in feature-based representations. These representations present the advantage of being translatable into other conventional formats~\cite{gu2021stylenerf,epigraf,EG3D}.
The majority of 3D modeling methods have been grounded in direct supervision, leveraging labeled 3D attribute ground truth for enhanced accuracy~\cite{peng2020convolutional, wang2021locally, liu2021deep}. However, such methods pose challenges in real-world applications, primarily due to the resource-intensive nature of labeling 3D ground truth. This often involves intricate data collection and curation processes. 
In light of these challenges, unsupervised techniques, specifically 3D Generative Adversarial Networks (3D-GANs), have been introduced~\cite{unsuper-multi-view, unsupervised-recon-realistic, cha2019unsupervised-recon}. Yet, their potential in 3D modeling often falls short of optimal performance.

In a bid to bridge this performance gap, self-supervised learning approaches have been proposed. These techniques capitalize on the inherent consistency between the 3D space and other domains, notably the 3D-2D consistency~\cite{canonical-mesh-uv, Hu_2021_CVPR, multi-view-consistency, self-UV-map}. A remarkable example is the work by \cite{Hu_2021_CVPR}, which leveraged the principles of CycleGAN~\cite{CycleGAN2017} to establish landmark consistency and interpolated consistency, ultimately facilitating the learning of 3D mesh representations. Nonetheless, there remains a significant research gap in effectively integrating Self-Supervised Learning (SSL) with 3D-GANs.

\noindent\textbf{3D-GANs.} In recent years, the remarkable success of 2D-GANs in high-resolution image synthesis~\cite{stargan-AFHQ, StyleGAN3D} has paved the way for the emergence of 3D-GANs~\cite{gu2021stylenerf, chan2021pi, schwarz2020graf, GET3D, EG3D, or2022stylesdf, henderson20cvpr, 2DGANSMeet, dynamic_Discriminator, meet_the_2D3D, karmali2022hierarchical, triangle, mueller2022diffrf}. While various 3D feature representations are used in these approaches, 3D voxel~\cite{gu2021stylenerf, Structural_Textural_Representation, or2022stylesdf} and more efficient alternatives like triplanes~\cite{EG3D} have gained prominence due to memory and computational considerations. Following representation acquisition, 3D-GANs employ NeRF~\cite{nerf} for rendering, and additional 2D layers refine rendered maps to yield photorealistic images.
Nonetheless, the quality of 3D generation from the obtained representations remains constrained, primarily due to a focus on 2D domain supervision while overlooking practical 3D constraints~\cite{waibel2022diffusion}. Consequently, existing approaches often yield 3D generation outcomes with irregular surface shapes, notably hole artifacts. Although prior works have introduced constraints for density and signed distance values, their impact on enhancing 3D modeling realism is limited, largely acting as regularization terms.

In response, we propose effective 3D constraints to train 3D-GANs without necessitating explicit 3D annotations. Our strategy serves as a plug-and-play module, enhancing the 3D generation outcomes of generative networks across diverse architectures.

\vspace{-.1in}
\section{Methodology}

\subsection{Notations}
As shown in Fig.~\ref{fig:framework}, in the realm of 3D-GANs, the standard architecture comprises a latent space $\mathcal{Z}$, a 3D representation generator $\mathcal{G}$, a geometrical formulator $\mathcal{M}$, and a 2D rendering module $\mathcal{R}$. Given a latent variable $z\in \mathcal{Z}$, the associated 3D representation $r$ is computed as $r=\mathcal{G}(z)$. Regardless of the structural intricacies, such as triplane geometries~\cite{EG3D}, $r$ can be tailored to a unified voxel representation. The geometric modeling process, e.g., mesh and point cloud generation, transforms $r$ using the procedure in $\mathcal{M}$, resulting in $m=\mathcal{M}(r)$. For rendering outputs, the NeRF-based rendering module $\mathcal{R}$ leverages $r$ and the target camera pose $p$ to yield $I=\mathcal{R}(r, p)$.
In this study, our focal point is the integration of SSL constraints onto $r$, with a particular emphasis on refining geometrical modeling. Simultaneously, enhancing the quality of $r$ bears the potential to augment the overall rendering outcomes $I$.

\subsection{Cyclic Generative Constraint (CGC)} \label{sec:cycle}

\noindent\textbf{Motivation.} 
We introduce a cyclic generative constraint aimed at expanding the discrete valid area within the latent space of 3D-GANs. Our approach is motivated by the desire to extend a given latent point $z$ to corresponding local points $z^*$. These local points $z^*$ share the same geometrical representation as $z$, and the key idea is to enforce geometric outputs generated from $z$ and $z^*$ to match. By doing so, we mitigate abrupt shifts in the latent space, thereby enhancing its density and smoothness. To achieve local augmentation, we leverage a 3D-GAN inversion technique that employs our custom-designed transformer-based encoder $\mathcal{E}$, the specifics of which will be detailed in the forthcoming section.

\vspace{1mm}
\noindent\textbf{Differences with existing works.} 
Our proposed cyclic self-training constraint (CGC) differs from the cycle consistency loss utilized in unpaired 2D translation methods~\cite{CycleGAN2017,xu2019view,hoffman2018cycada}. In the 2D context, the cycle training constraint relies on the assumption of bijections at the image level, aiming to disentangle attributes for unpaired translation. In contrast, our approach utilizes a cyclic path between 3D features and latent features to establish a dense and smooth distribution within the latent feature space. This, in turn, facilitates the generation of valid 3D representations and mitigates the risk of abnormal mappings from the latent space to 3D representation.
Furthermore, our CGC also stands apart from unsupervised objectives in 2D-3D translation~\cite{Hu_2021_CVPR}. In those cases, useful information may be lost during the 3D-to-2D differential projection process, whereas our approach retains complete geometrical knowledge within both the 3D representation and latent feature. The well-balanced information present in our cycle constraint proves beneficial for distribution adjustment.

\vspace{1mm}
\noindent\textbf{Implementation: data flow.} 
To implement CGC (Fig.~\ref{fig:framework}), our process begins by randomly sampling a latent variable $z$ from the latent space $\mathcal{Z}$ using a Gaussian distribution. This sampled $z$ is then mapped to a 3D representation denoted as $r$. To obtain the corresponding local points, we employ an inversion procedure on $r$, resulting in a latent point $z^*$ through a transformer-based encoder $\mathcal{E}$. This $z^*$, in turn, leads to the generation of another 3D representation, designated as $r^*$. The encoder $\mathcal{E}$ is trained alternately with the generator and discriminator of the 3D-GANs, ensuring a consistent local relation between $z$ and $z^*$. Subsequently, CGC is applied to the 3D representations $r$ and $r^*$.

In summary, the data flow in our framework's training stage involves the path of generation-inversion-generation, and can be summarized as the following equation, as

\begin{equation}
\small
\begin{aligned}
    &r=\mathcal{G}(z), z^* = \mathcal{E}(r), r^*=\mathcal{G}(z^*), \\
    &z/z^* \in \mathbb{R}^{C_z}, r/r^* \in  \mathbb{R}^{S\times S \times S \times C_r},
\end{aligned}
    \label{eq:cycle}
\end{equation}
where $C_z$ is the feature channel of the latent space, $C_r$ is the feature channel of the voxel-based 3D representation, and $S$ is the corresponding size.

\begin{figure}
    \centering
    \centerline{\includegraphics[width=1\linewidth]{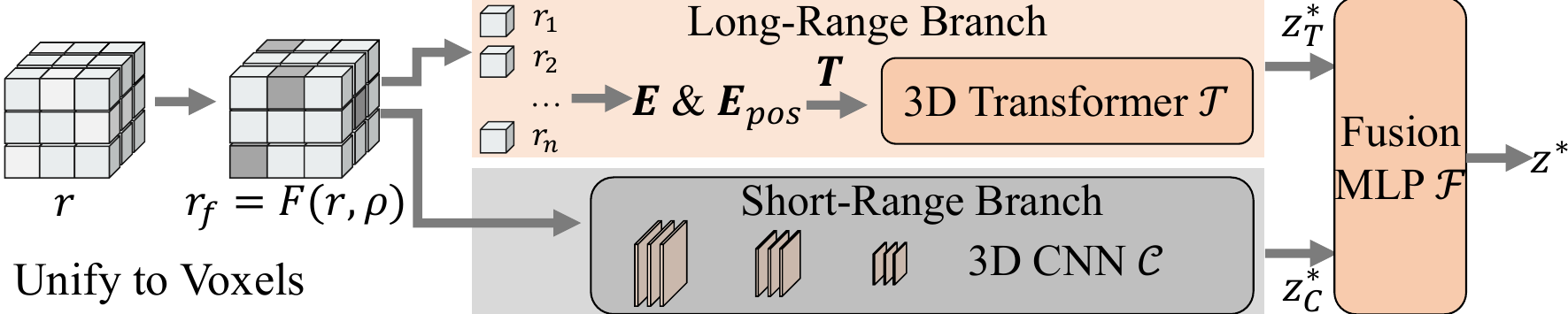}}
    \vspace{-0.1in}
    \caption{
   \small The detailed implementation of the encoder in CGC. The inversion encoder for 3D-GANs combines CNN and transformer components, operating in tandem to incorporate both short- and long-range modeling. This design ensures the preservation of geometrical information between $z$ and $z^*$, as depicted in Fig.~\ref{fig:framework}.
    }
    \label{fig:framework-encoder}
    \vspace{-0.25in}
\end{figure}

\vspace{1mm}
\noindent\textbf{Implementation: loss function.} 
In our framework, we initially train the encoding module $\mathcal{E}$ for a set of warm-up epochs, which has been observed to yield benefits for the joint inversion and generation process.
The input to $\mathcal{E}$ is standardized to be a voxel-based representation, which can be derived from various other 3D representations (for more details, please refer to supp). It is important to note that the input to $\mathcal{E}$ can be adapted to different formats, and in this paper, we use voxels as an illustrative example, given its common usage.
$\mathcal{E}$ can be trained with the objective as
\begin{equation}
\small
    \mathcal{L}_{Z} = \sum_{z\in \mathcal{Z}} \| z - z^* \|_1.
    \label{eq:gaussian_c}
\end{equation}

Following the warm-up training of $\mathcal{E}$, the generator $\mathcal{G}$ and discriminator $\mathcal{D}$ of the 3D-GANs can be alternatively trained in conjunction with $\mathcal{E}$.
Given the paired 3D representations $r$ and $r^*$ obtained as per Eq.~\ref{eq:cycle}, our approach optimizes the distance between them to realize our intended objective. This optimization leverages the proximate distance between $z$ and $z^*$ facilitated by $\mathcal{E}$.
Thus, the constraint for the 3D representation can be written as
\begin{equation}
\small
    \mathcal{L}_{R} = \sum_{z\in \mathcal{Z}} \| r - r^* \|_1.
    \label{eq:gaussian_r}
\end{equation}
Empirical evidence demonstrates that CGC effectively enhances the sharpness and realism of geometrical modeling in 3D-GANs, mitigating diverse artifacts, as illustrated in the experiments section. The convergence of losses in both latent space ($\mathcal{L}{Z}$) and 3D representation space ($\mathcal{L}{R}$) is also visually depicted therein.

\subsection{Encoder Architecture}
\label{sec:encoder}

The quality of inversion $\mathcal{E}$ significantly impacts the effectiveness of CGC, with a primary requirement being the \textit{retention of geometrical information during inversion}.
Conventionally, GAN inversion employs Convolutional Neural Network (CNN) models as encoders. However, CNNs struggle with capturing non-local 3D spatial information, essential for detailed geometrical modeling that demands a balance of local and global features.
In contrast, transformers have recently proven adept at capturing non-local features effectively~\cite{vaswani2017attention,liu2021swin,xu2022snr}.
Hence, we introduce a novel encoder structure for inverting 3D representations into the latent space. This design amalgamates the strengths of transformers and CNNs (Fig.~\ref{fig:framework-encoder}), enabling adaptive spatial-varying operations.

\vspace{1mm}
\noindent\textbf{The input filtering of the encoder.} 
We've identified that the inversion quality can be affected by regions with low density. Often, these regions consist of empty spaces that are erroneously interpreted as occupied due to their low density.
Thus, we add an adaptive filter operation for the input $r$, as
\begin{equation}
\small
    r_f = F(r ; \rho),
\end{equation}
where $F$ is the filter operation, i.e., change the value of the target area to be zero whose density is lower than $\rho$. $\rho$ is a learnable parameter, which is randomly initialized by an empirical value.

\vspace{1mm}
\noindent\textbf{Overview: the structure of the encoder.} 
The encoder can be partitioned into three components: a long-range branch using a transformer, a short-range branch utilizing a CNN, and an MLP head responsible for fusion and generating the output latent variable.
Thus, the encoding process in Eq.~\ref{eq:cycle} can be written as
\begin{equation}
\small
\begin{aligned}
    &z_T^*=\mathcal{T}(r_f), \,  z_C^*=\mathcal{C}(r_f), \, z^*=\mathcal{E}(r)=\mathcal{F}(z_T^*\oplus z_C^*),
\end{aligned}
\label{eq:sl}
\end{equation}
where $\mathcal{T}$ denotes the transformer, $\mathcal{C}$ is the CNN part, and $\mathcal{F}$ denotes the MLP, $\oplus$ means the concatenation operation.

\vspace{1mm}
\noindent\textbf{The input tokenization for $\mathcal{T}$.} 
Similar to the 2D vision transformer~\cite{vaswani2017attention}, the input of the 3D transformer $\mathcal{T}$ are the voxel-based tokens with the corresponding position information. Suppose the size of 3D representation is $S\times S \times S$, and the voxel size of each token is $p\times p \times p$. The total number of voxel-based tokens for $r_f$ is $n=(\frac{S}{p})^3$, as $\{ r_1, r_2, \dots , r_n \}$. The final input feature tokens of the transformer can be obtained via a projection matrix $\textbf{E}$ and the corresponding position embedding matrix $\textbf{E}_{pos}$, as 

\begin{equation}
\small
\begin{aligned}
&\mathbf{T}=\left[r_1 \mathbf{E} ; r_2 \mathbf{E} ; \cdots ; r_n \mathbf{E}\right]+\mathbf{E}_{pos},\\
&  r_i\in \mathbb{R}^{p^3 \cdot C_r},  \mathbf{E} \in \mathbb{R}^{\left(P^3 \cdot C_r\right) \times D}, \mathbf{E}_{pos} \in \mathbb{R}^{n \times D},
\end{aligned}
\end{equation}
where $D$ is the channel of final input feature tokens.
$\mathbf{T}$ is further processed by the transformer $\mathcal{T}$ (Eq.~\ref{eq:sl}).

\subsection{Overall Training}
\label{sec:overall}
Our SSL loss functions are applicable to all existing 3D-GANs. Our proposed Cyclic Generative Constraint is not contradictory with the original loss for the generator, e.g., 2D GAN loss. Thus, the overall loss for the generator $\mathcal{G}$ is 
\begin{equation}
\small
    \mathcal{L}_G=\mathcal{L}_{GAN, G}+\lambda \mathcal{L}_{R},
\end{equation}
where $\mathcal{L}_{GAN, G}$ is the 2D GAN loss, $\lambda$ is the corresponding loss weight.
The loss for the discriminator is kept as the original discriminate loss $\mathcal{L}_{GAN, D}$. Moreover, the structures of the original generator and discriminator are not needed to be modified. We only need to add one encoding module $\mathcal{E}$ into the training with its loss as $\mathcal{L}_{Z}$.
Extensive experiments show that our proposed SSL strategy can be employed as a plug-and-play module in existing 3D-GANs, enhancing their geometrical modeling results and 2D rendering performance.

\begin{figure*}
    
    \centering
    \centerline{\includegraphics[width=1\linewidth]{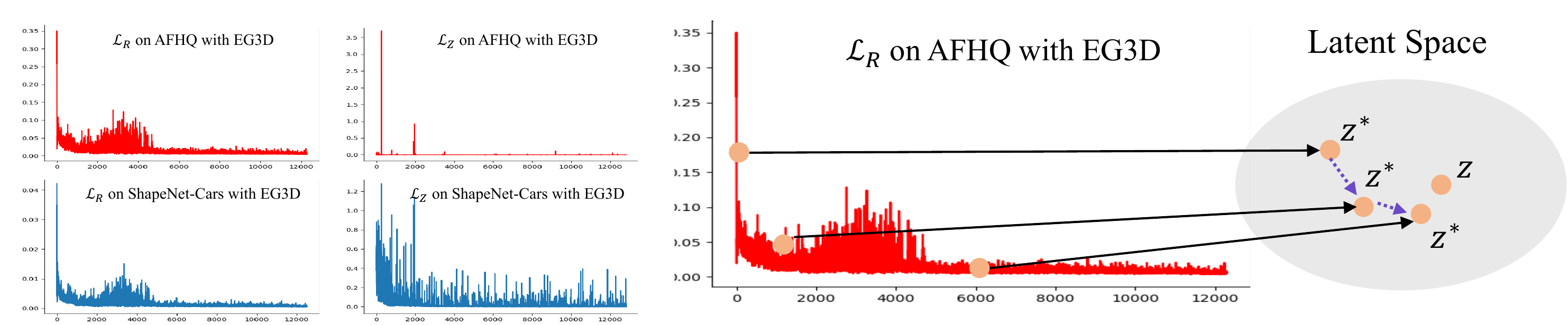}}
   \vspace{-0.1in}
   \caption{\small (Left) These figures show the convergence of $\mathcal{L}_Z$/$\mathcal{L}_R$ on two datasets, i.e., AFHQ and ShapeNet-Cars. 
   It's demonstrated that $\mathcal{L}_Z$ during training will be fast converged, guaranteeing the correct local relation between $z$ and $z^*$ in Eq.~\ref{eq:cycle}. 
   (Right) our encoder is not over-fitted during the training since the MSE loss is not converged to zero. This also implies the change of $z^*$: it moves towards $z$ from far to near, i.e., traversing the local area of $z$ to find sufficient satisfied $z^*$.
   }
   \vspace{-0.1in}
    \label{fig:MSE}
\end{figure*}

\begin{table*}[]
\centering
\resizebox{1.0\linewidth}{!}{
\begin{tabular}{l|cc|cc|cc}
\toprule
Experiment                     & FFHQ-FID & FFHQ-KID & AFHQ-FID & AFHQ-KID & ShapeNet-Cars FID & ShapeNet-Cars KID    \\ \hline
StyleSDF~\cite{or2022stylesdf} & 12.0     & 4.19     & 10.3     & 6.90     &     --      &         --               \\                        
StyleSDF+Ours                  &\textbf{9.9}    &    \textbf{3.37}      & \textbf{8.3}    &  \textbf{4.77}   & --   &      --                            \\ \hline
EG3D~\cite{EG3D}               & 5.2      & 2.85     & 8.3     & 0.47     & 4.23     &   0.210                  \\             
EG3D+Ours                      &\textbf{5.0}   &     \textbf{2.75}    & \textbf{7.3}     &     \textbf{0.24}     & \textbf{3.96 }      &  \textbf{0.170}                    \\              
\bottomrule
\end{tabular}}
\vspace{-0.1in}
\caption{
\small Comparison between baselines and methods with our SSL. 
Experiments are conducted on three representative datasets.
}
\label{table:baseline}
\vspace{-0.2in}
\end{table*}

\section{Experiments}

\subsection{Datasets}
Experiments encompasses three datasets: \textbf{FFHQ}~\cite{kazemi2014one}, \textbf{AFHQ}~\cite{stargan-AFHQ}, and \textbf{ShapeNet}~\cite{chang2015shapenet} dataset which includes images of cars and chairs. FFHQ comprises 70,000 human face images, while AFHQ includes 15,630 animal face images. For fairness, we adhere to existing 3D-GANs' training and testing split settings, such as EG3D~\cite{EG3D}. Additionally, we leverage the ShapeNet dataset for evaluation, which comes with 3D geometry annotations, allowing us to assess the geometric improvements achieved through our SSL strategy. Specifically, the evaluation centers on the "cars" and "chairs" categories within ShapeNet. The geometric evaluation employs a conditional generation setup with input images, following the standard split scheme in ShapeNet.

\subsection{Details} 

\noindent\textbf{Baselines and implementation details.} 
Our SSL strategy is universally applicable to existing 3D-GANs. For experiments, we select representative models, like StyleSDF~\cite{or2022stylesdf}, EG3D~\cite{EG3D}, and GET3D~\cite{GET3D}. Implementation of our SSL strategy builds upon their officially released codebases. Our adaptations entail introducing the encoder $\mathcal{E}$, along with $\mathcal{L}{R}$ and $\mathcal{L}{Z}$. Training was carried out using 8 NVIDIA RTX 3090 GPUs.

\vspace{1mm}
\noindent\textbf{Details of the encoder.}
The encoder comprises both a transformer and a CNN component, implemented as a 3D-Res-UNet~\cite{zgniek20163DUL} and 3D voxel transformers~\cite{YimingLi2023VoxFormerSV}, respectively. Input voxel sizes are configured as $64^3$ (can also be varying for different frameworks). 

\subsection{Metric}
\subsubsection{Metric for 2D generation.}
We follow current works, and use Fréchet Inception Distance scores~\cite{FID} (\textbf{FID}) and Kernel Inception Distance~\cite{binkowski2018demystifying} (\textbf{KID}) as our evaluation metric. For FID and KID, the lower, the better. For KID, we report its value $\times 100$.

\subsubsection{Metric for 3D geometry generation.}
Besides the standard quantitative evaluation of 2D rendering results, we propose the assessment of 3D geometrical information with the metric of ``3D IoU" and ``3D discriminative score".

\noindent\textbf{1) 3D IoU.}
For datasets with 3D annotations, the generator can be set as the conditional setting with the input of an image besides $z$ (i.e., employ the generator for single-view-based 3D reconstruction), and we verify whether the generated 3D representation is aligned with the ground truth. 
We first change both $r$ and the annotation into voxels and then employ 3D IoU for evaluation. The higher the better, which reflects the accuracy and realism of the reconstruction.

\noindent\textbf{2) 3D Discriminative Score (3D DS)}
To compute the 3D discriminative score, we initiate by training a classifier. This classifier takes the synthesized or real 3D mesh $m$ as input and provides the corresponding object category as output. Our choice of classifier is PolyNet~\cite{3D-poly}, known for its high accuracy and efficiency in 3D object classification tasks. For training purposes, we utilize the ModelNet-10 dataset~\cite{wu20153d}, containing 4899 3D models across 10 categories. The trained classifier achieves a real 3D object accuracy of 92\%.
In evaluating the performance of the generator $\mathcal{G}$, we generate $B$ 3D meshes with a specified category label $l$. Subsequently, if the trained classifier predicts $C$ meshes as category $l$, the 3D discriminative score is computed as $\frac{C}{B}$. This score quantifies the perceptual quality of the generated 3D objects. 

\subsection{Consistent Convergence for $\mathcal{L}_{R}$ and $\mathcal{L}_{Z}$}
\label{sec:convergence}
As shown in Fig.~\ref{fig:MSE}, we visualize the value of $\mathcal{L}_{Z}$ and $\mathcal{L}_{R}$ with respect to the training step. This figure demonstrates that loss convergence in both latent space and 3D representation space is achieved. Specifically, $\mathcal{L}_{Z}$ is converged to a small value in the early epoch of the training, guaranteeing the locality between $z$ and $z^*$ in CGC optimization. Moreover, our encoder is not overfitted during the training since the MSE loss is not converged to zero. This also implies the change of $z^*$: \textbf{it moves towards $z$ from far to near}, i.e., traversing the local area of $z$ to find sufficient satisfied $z^*$.

\begin{table}[t]
\centering
\resizebox{1.0\linewidth}{!}{
\begin{tabular}{l|p{5cm}<{\centering}}
\toprule
Experiment    & 3D IoU on ShapeNet-Cars   \\ \hline
EG3D~\cite{EG3D}                        &    47.2     \\
EG3D+Ours                               &    \textbf{52.2}     \\ \hline
GET3D~\cite{GET3D}                      &    56.1         \\
GET3D+Ours                              &     \textbf{57.4}        \\ 
\bottomrule
\end{tabular}}
\vspace{-0.1in}
\caption{\small Comparison between baselines and methods with our SSL, in terms of 3D IoU results.}
\label{table:3DIOU}
\vspace{-0.15in}
\end{table}

\begin{table}[t]
\centering
\resizebox{1.0\linewidth}{!}{
\begin{tabular}{l|p{2.5cm}<{\centering}|p{2.5cm}<{\centering}}
\toprule
Experiment & ShapeNet-Cars & ShapeNet-Chairs \\ \hline
EG3D~\cite{EG3D}      &         88.2       & ---             \\
EG3D+Ours  &       \textbf{92.0 }       & ---             \\ \hline
GET3D~\cite{GET3D}      &    95.2           &    80.0             \\
GET3D+Ours &       \textbf{95.8}        &               \textbf{85.2} \\
\bottomrule
\end{tabular}} 
\vspace{-0.1in}
\caption{\small Comparison between baselines and methods with our SSL, in terms of 3D DS results.}
\label{table:3D-Dis}
\vspace{-0.3in}
\end{table}

\begin{table}[t]

    \centering
 \resizebox{1.0\linewidth}{!}{
\begin{tabular}{l|p{5cm}<{\centering}}   
    \toprule
    Experiment                 & FID           \\ \hline
    Baseline         & 8.30 \\ \hline
    wo 3D-Transformer &   7.20                 \\
    wo 3D-CNN         &    8.71             \\
    \hline 
    Ours             &     \textbf{7.10}  \\
    \bottomrule
    \end{tabular}}
    \vspace{-0.1in}
    \caption{\small The ablation study with different structures for $\mathcal{E}$. Experiments are conducted on AFHQ~\cite{stargan-AFHQ} with EG3D~\cite{EG3D}.}
    \label{tab:abla_encoder}
    \vspace{-0.1in}
\end{table} 

\begin{figure*}[ht]
    \centering
    \centerline{\includegraphics[width=1.0\linewidth]{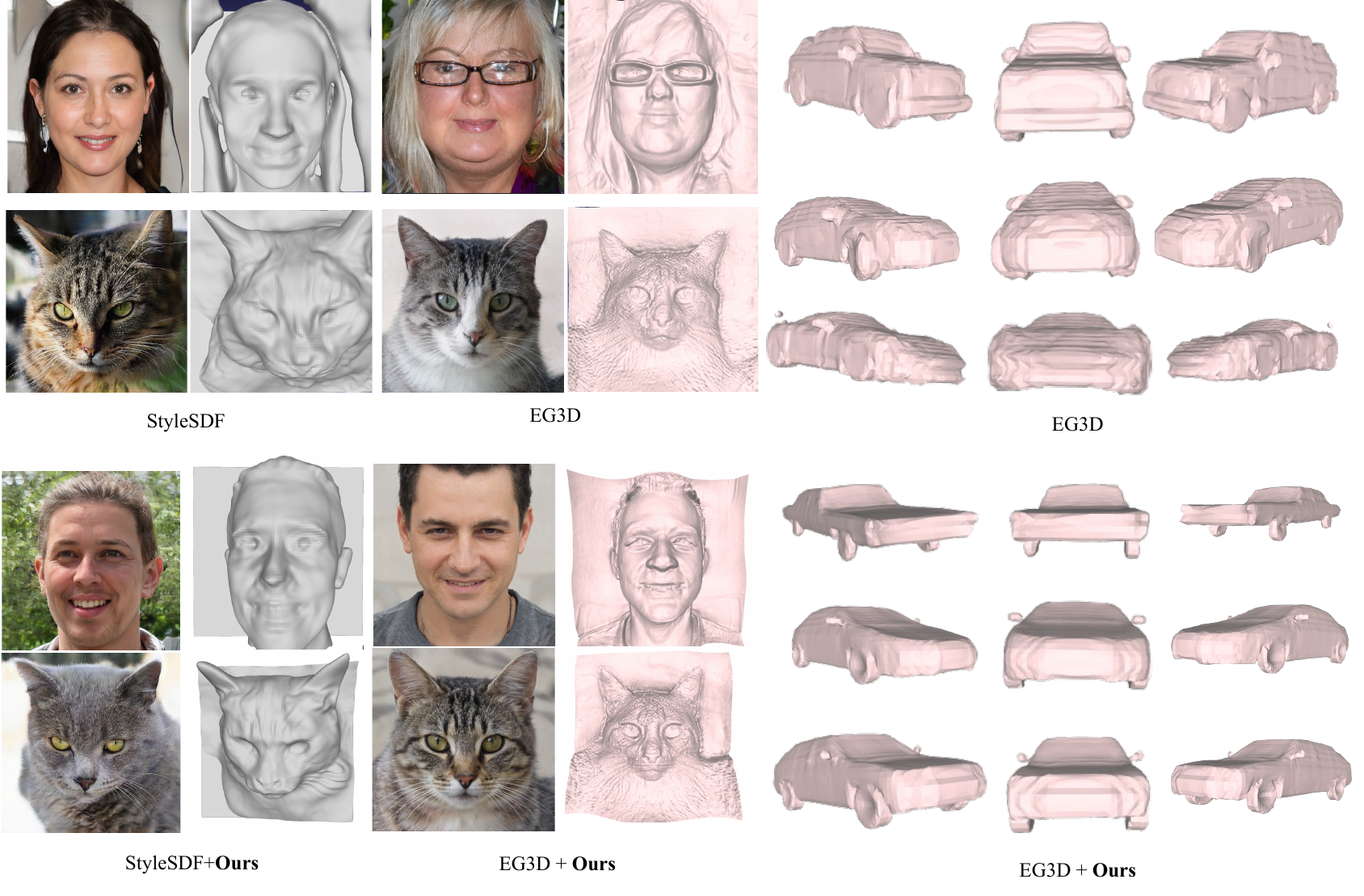}}
    \vspace{-0.2in}
    \caption{\small The visual comparison between baselines and methods with our SSL. Our geometrical modeling is more realistic with fewer artifacts and wrong surfaces.}
    \label{fig:compare}
   \vspace{-0.2in}
\end{figure*}

\begin{figure}[t]
    \centering
    \centerline{\includegraphics[width=1.0\linewidth]{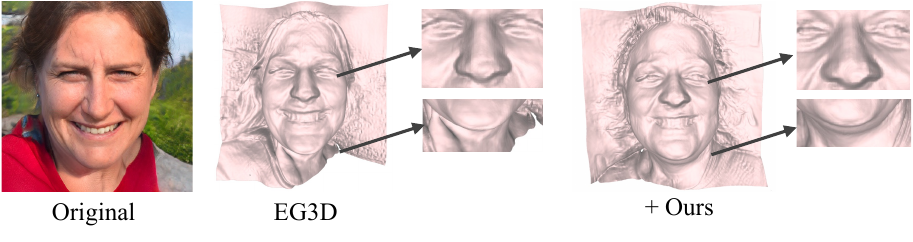}}

    \vspace{-.2in}
    \caption{\small The 3D-aware geometry comparison with the pivot inversion which inverses the original real-world image into the latent space. As shown, our geometrical result (right) is better than the baseline (left), when we inverse the same image into the latent.}
    \label{fig:PTI}
    \vspace{-.2in}
\end{figure}

\subsection{Quantitative Comparison}

\noindent\textbf{2D image generation comparison.}
Following current 3D-GANs, we use FID and KID as the metric to evaluate the 2D rendering results.
When computing FID and KID, we adopt the original full resolution for images from different datasets, i.e., 1024 for FFHQ, 512 for AFHQ, 128 for ShapeNet (more details can be found in supp.).
As shown in Table~\ref{table:baseline}, the models equipped with our SSL strategy will have improvements in 2D rendering results.
The improvement of rendering effects is derived from the perfection of 3D generation since it provides effective physical priors for rendering and 2D synthesis.

\vspace{1mm}
\noindent\textbf{3D generation comparison with 3D IoU.} 
The results of the 3D IoU comparisons are displayed in Table \ref{table:3DIOU}. These comparisons demonstrate that the reconstructed 3D shapes with our strategy are better aligned with ground truths. 
Thus, 3D-GANs trained with our SSL strategy can lead to more reliable and accurate results.

\vspace{1mm}
\noindent\textbf{3D generation comparison with 3D DS.}
3D DS comparison results are presented in Table \ref{table:3D-Dis}.
It's shown that our method outperforms the baseline by a significant margin, indicating that the methods combined with our SSL strategy generate more accurate and realistic 3D objects.

\subsection{Qualitative Comparison} 
We compare the visual quality of our 3D generation with the baselines, as shown in Fig.~\ref{fig:compare}. 
We can see that our SSL constraints enhance the results of both StyleSDF and EG3D, showing that our performance is not dependent on the architecture of 3D-GANs.
For example, better details can be observed in the results of StyleSDF combined with our constraints. Besides the sharp visual effects, our generation results have fewer irregular surfaces.
Compared with the original EG3D, although both methods contained the physically valid structure, ours can produce a more realistic 3D surface. For example, on the generated cat ears, the results with our constraints have a more detailed shape. 
Moreover, our method has prominent geometrical superiority compared with the baseline when compared on the ShapeNet-cars.

\begin{table}[t]
\centering
\resizebox{1.0\linewidth}{!}{
\begin{tabular}{l|p{3.0cm}<{\centering}|p{3.0cm}<{\centering}}  
\toprule
Experiment                  & FID & KID \\ \hline 
$z+z^*$                     &  7.07   &  \textbf{0.21}  \\
$z+z^*+z^{**}$              &  \textbf{7.06}   &  0.23   \\
\bottomrule
\end{tabular}}
\vspace{-.1in}
\caption{\small Ablation studies to verify the integrity of geometrical information during inversion and generation. Experiments are conducted on AHFQ~\cite{stargan-AFHQ} with EG3D~\cite{EG3D}.}
\label{tab:multiple}
\end{table}

\vspace{-.1in}
\subsection{Ablation Study}

\begin{table}[t]
\centering
\vspace{-.1in}
 \resizebox{1.0\linewidth}{!}{
\begin{tabular}{l|c|c|c|c}
\toprule
Datasets      &  GIRAFFE  & +Laplacian Loss & +Local Search & +Ours \\ \hline
Cat-64   & 8  & 10             &  8            &  \textbf{7}            \\
Car-64   & 16 & 17             &  14            & \textbf{13}             \\
Chair-64 & 20 & 25             &  --            & \textbf{18}             \\ \hline
Cat-256  & 19 & 23             &  --            &  \textbf{16}            \\
\bottomrule
\end{tabular}}
\vspace{-0.1in}
\caption{\small Ablation study on the self-supervised alternative methods, by using ShapeNet and AFHQ to compute FID score.}
\label{ablation_ssl}
\vspace{-0.2in}
\end{table}

\noindent\textbf{Different structures for encoders.}
Different from existing 3D-GANs inversion approaches, we design a new strategy with input of 3D representations, to avoid the loss of geometrical information. Specifically, we design a novel encoder strategy with spatially combined CNN and transformer features. We set an ablation study to show the effectiveness of our designed encoder, by deleting CNN and transformer respectively. As shown in Table \ref{tab:abla_encoder}, the results with only CNN and transformer structures are weaker than ours since it's hard for them to simultaneously capture both short- and long-range information.

\begin{figure}[t]
    \centering
    \centerline{\includegraphics[width=0.5\textwidth]{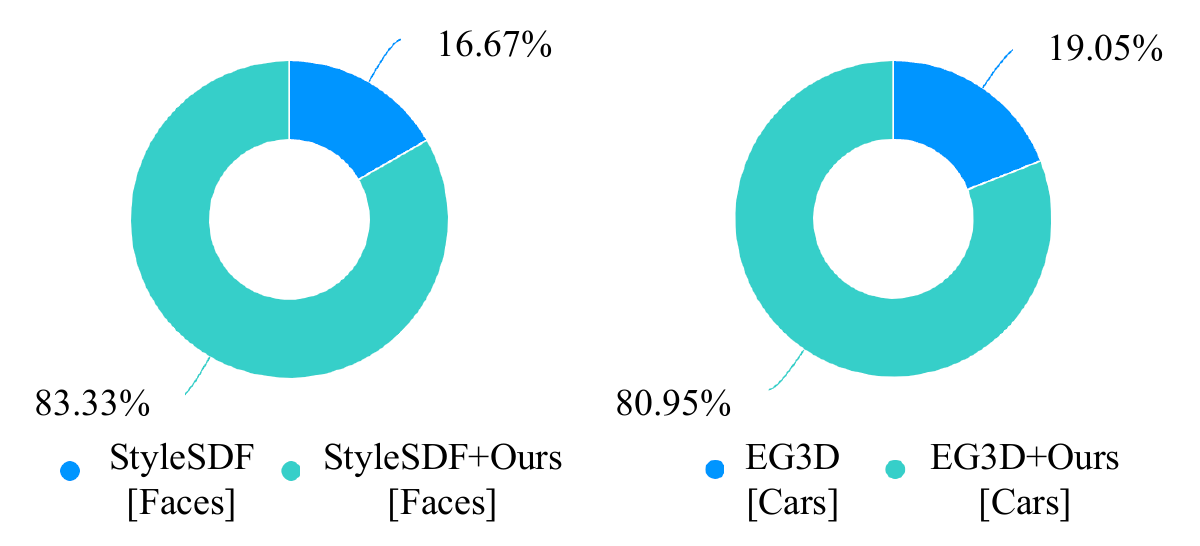}}
    \vspace{-0.15in}
    \caption{\small The user study result based on two datasets (faces~\cite{stargan-AFHQ, style-FFHQ} and ShapeNet~\cite{chang2015shapenet}) and network structures.}
    \label{fig:user}
    \vspace{-0.2in}
\end{figure}

\vspace{1mm}
\noindent\textbf{Integrity of geometrical information with our inversion.}
We set experiments to validate that our inversion strategy can preserve the geometrical information. The idea is to conduct the inversion-generation process repeatedly, and the error in the latent space will be accumulated if the information is dropped.
For example, we can obtain $z, z^*, z^{**}$ and $r, r^*, r^{**}$ by conducting Eq.~\ref{eq:cycle} twice, and apply Eqs.~\ref{eq:gaussian_c} and~\ref{eq:gaussian_r} for them. 
If the information is dropped, then the geometrical information among $z, z^*, z^{**}$ should be inconsistent, joint training with them will cause the performance drop of our approach.
The results are listed in Table~\ref{tab:multiple}. 
As shown, the results are almost the same, demonstrating that the geometrical information is preserved in our inversion strategy.

\vspace{1mm}
\noindent\textbf{Pivot Tuning Inversion comparison} 
In Fig.~\ref{fig:PTI}, we present the outcomes of inverting testing images using two distinct generators to obtain their respective 3D geometry representations. Our assessment employs the PTI~\cite{roich2022pivotal} technique on the FFHQ dataset. As depicted in Fig.~\ref{fig:PTI}, the geometrical result produced by our model (right) surpasses the baseline (left) when inverting the same image into the latent space. This observation underscores the superior capability of our model, guided by our strategy, to capture 3D modeling nuances for real-world images.

\vspace{1mm}
\noindent\textbf{Comparison with other SSL alternatives.} 
There are some other SSL alternatives that are referred from existing 3D reconstruction approaches, e.g., minimizing the signal's Laplacian (``+Laplacian Loss") which is usually used in NERF as surface smoothness constraint, and constraining the randomly searched $z$ and $z^*$ (without the guarantee that they have the same geometrical information) match in 3D representation. We have conducted several experiments to prove the superiority of our method over them. The experiment is conducted on GIRAFFE~\cite{GIRAFFE}, which is a simple 3D-GAN framework to amplify the differences among different strategies' outcomes.
As shown in Table~\ref{ablation_ssl}, our results are better than these SSL alternatives.

\vspace{-.1in}
\section{User Study}
We conducted a user study to validate the perceptual quality of the generation results obtained with our SSL strategy through human assessment. The study was carried out on AFHQ, FFHQ, and ShapeNet dataset. To ensure objectivity, we randomly generated a series of 3D results from the model trained both with and without our SSL supervision.
In the user study, participants were presented with pairs for comparison. Each pair consisted of the baseline and our model's results. The order of presentation (left-right) was randomized to avoid bias. Participants were required to select the one that they perceived as better based on the quality of geometrical details and its consistency with the 2D rendering results. Each pair included three 2D images along with their corresponding 3D representations.
A total of 84 valid questionnaires were collected for analysis. The results of the user study, as depicted in Fig.~\ref{fig:user}, indicate a user preference for the results obtained with our SSL loss.

\vspace{-.1in}
\section{Conclusion}
The proposal introduces a new SSL approach to bolster the geometric representation in current 3D-GANs. By utilizing a cyclic generative constraint, it aims to enhance local smoothness and prevent abrupt changes in the 3D-GANs' latent space, thus creating a densely valid space. This SSL technique can be seamlessly integrated into various 3D-GANs with different 3D representations. Comprehensive tests on key datasets validate the efficacy of this method.

\noindent\textbf{Limitation and future work.}
As we have stated, although great improvement can be achieved by employing our SSL strategy, it causes extra training costs. How to reduce the cost to be negligible will be our future work. Moreover, more 3D generative structures, e.g., 3D-GANs and 3D-diffusion structures, and more types of datasets should be involved in experiments as future work.

\bibliography{main}

\clearpage

\appendix

\section{Appendix}

In this supplementary material, we provide a comprehensive exploration of our proposed method, diving deeper into our experimental results and the intricacies associated with them. Starting off, we detail the superior geometric generation quality that our approach is a relatively good method. Following this, we dedicate a section to visual samples that span both 2D image and 3D geometric generation results. Further, into the document, enthusiasts of metrics and evaluation methods will find a detailed discourse on the evaluation metrics of 3D DS and 3D IOU. Concluding the material, we turn our attention to the real-world applications of our research, discussing the sample set we employed for our user study questionnaire.

\begin{figure*}[t]
    \centering
    \centerline{\includegraphics[width=1\linewidth]{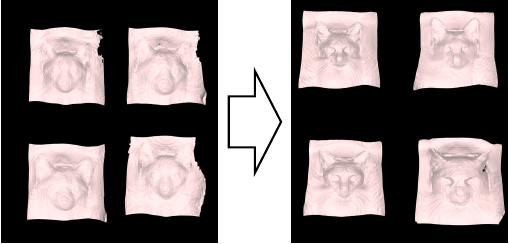}}
    \caption{
    The geometrical modeling results derived from the $64^3$ voxel-based 3D representation for the baseline and our method. Our approach has a shaper and more realistic shape, which verify the 3D representation employed in our approach has enough geometrical information to conduct CGC's inversion.
    }
    \label{fig:cat_1}
    
\end{figure*}

\begin{figure*}[t]
    \centering
    \centerline{\includegraphics[width=1.0\linewidth]{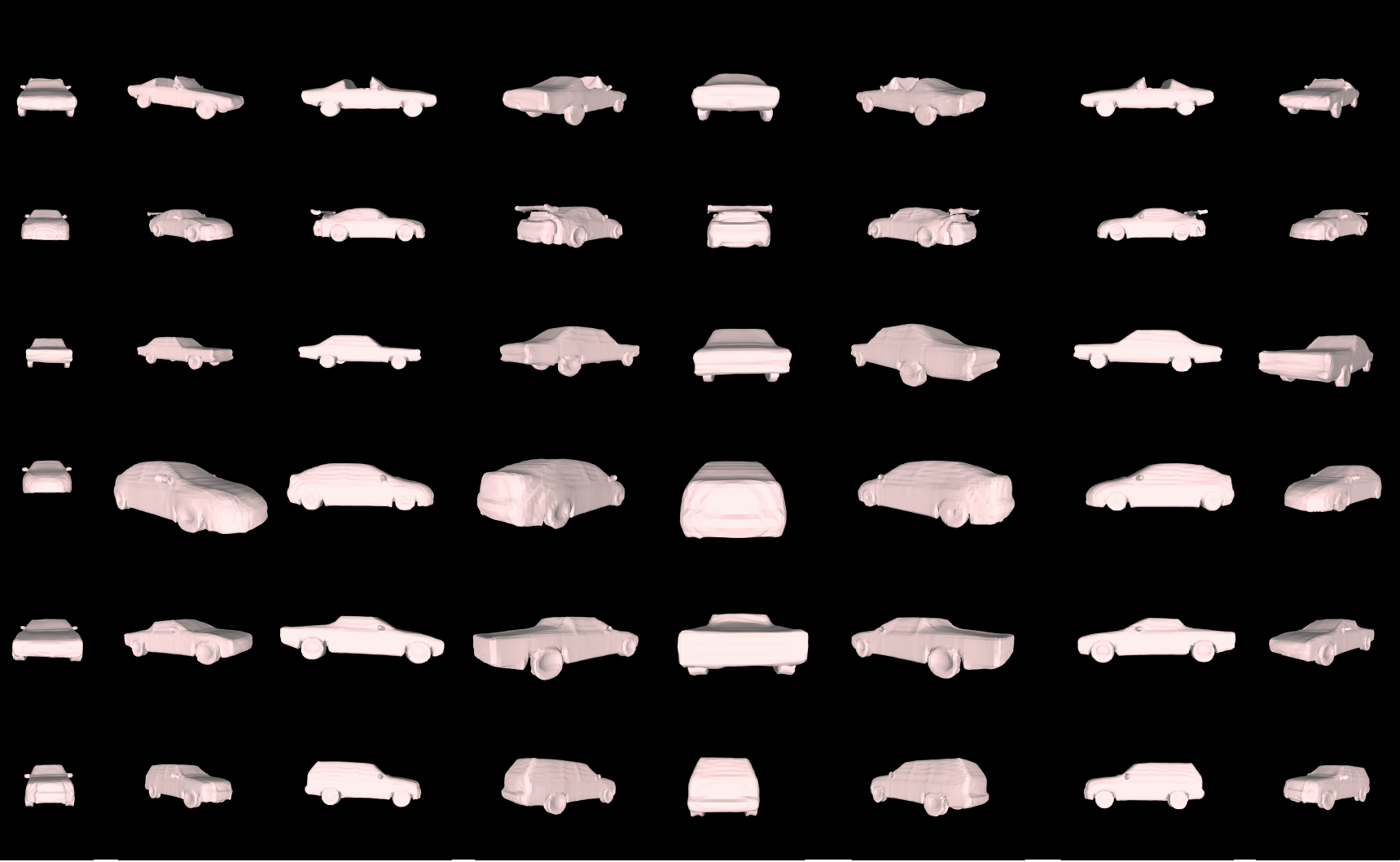}}
    \caption{The 3D visualization result on ShapeNet-Cars with EG3D+Ours.
    }
    \label{fig:car_1}
\end{figure*}

\vspace{-.1in}
\section{The Quality of Geometrical Modeling}
\label{sec:3d}

During the training phase, due to computational resource constraints, we employ the CGC using a low-resolution voxel size for the input 3D representation. The geometrical mesh derived from this low-resolution 3D representation is visualized for both baselines and our method in Fig.~\ref{fig:cat_1}. As evident from Fig.~\ref{fig:cat_1}, the geometric modeling enhanced by our SSL loss term exhibits increased sharpness and realism, underscoring the efficacy of our approach. Furthermore, the high quality of the geometric modeling attests to the adequacy of the 3D representation used in our method, proving it holds ample geometrical information for inversion.

\vspace{-.1in}
\section{Extended Visualization}
\label{sec:visual}

\begin{figure*}[t]
    \centering
    \centerline{\includegraphics[width=1.0\linewidth]{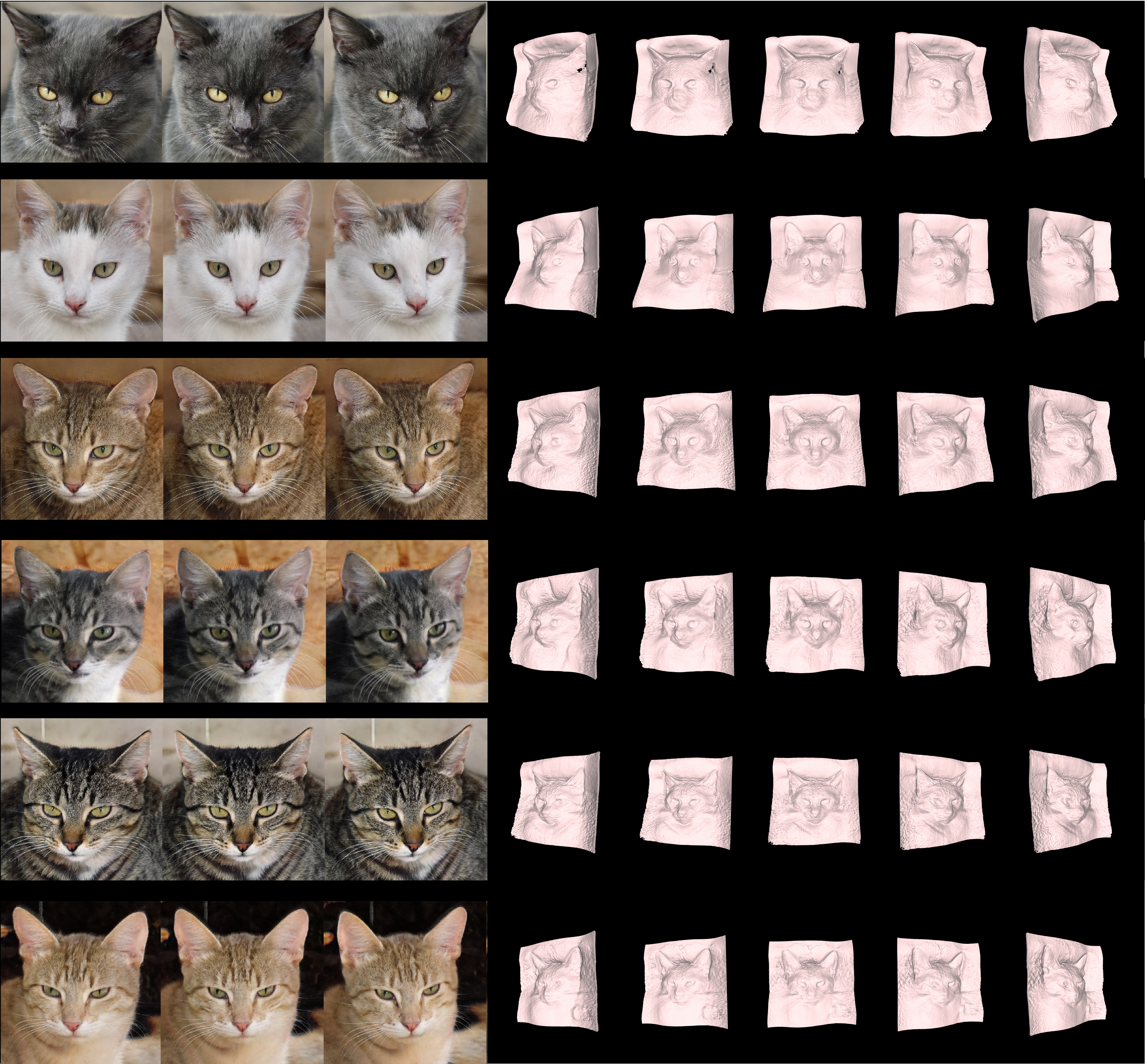}}
    \caption{The visualization for 2D and 3D generation result on the AFHQ dataset with EG3D+Ours.
    }
    \label{fig:cat_2}
\end{figure*}

\begin{figure*}[t]
    \centering
    \centerline{\includegraphics[width=0.9\linewidth]{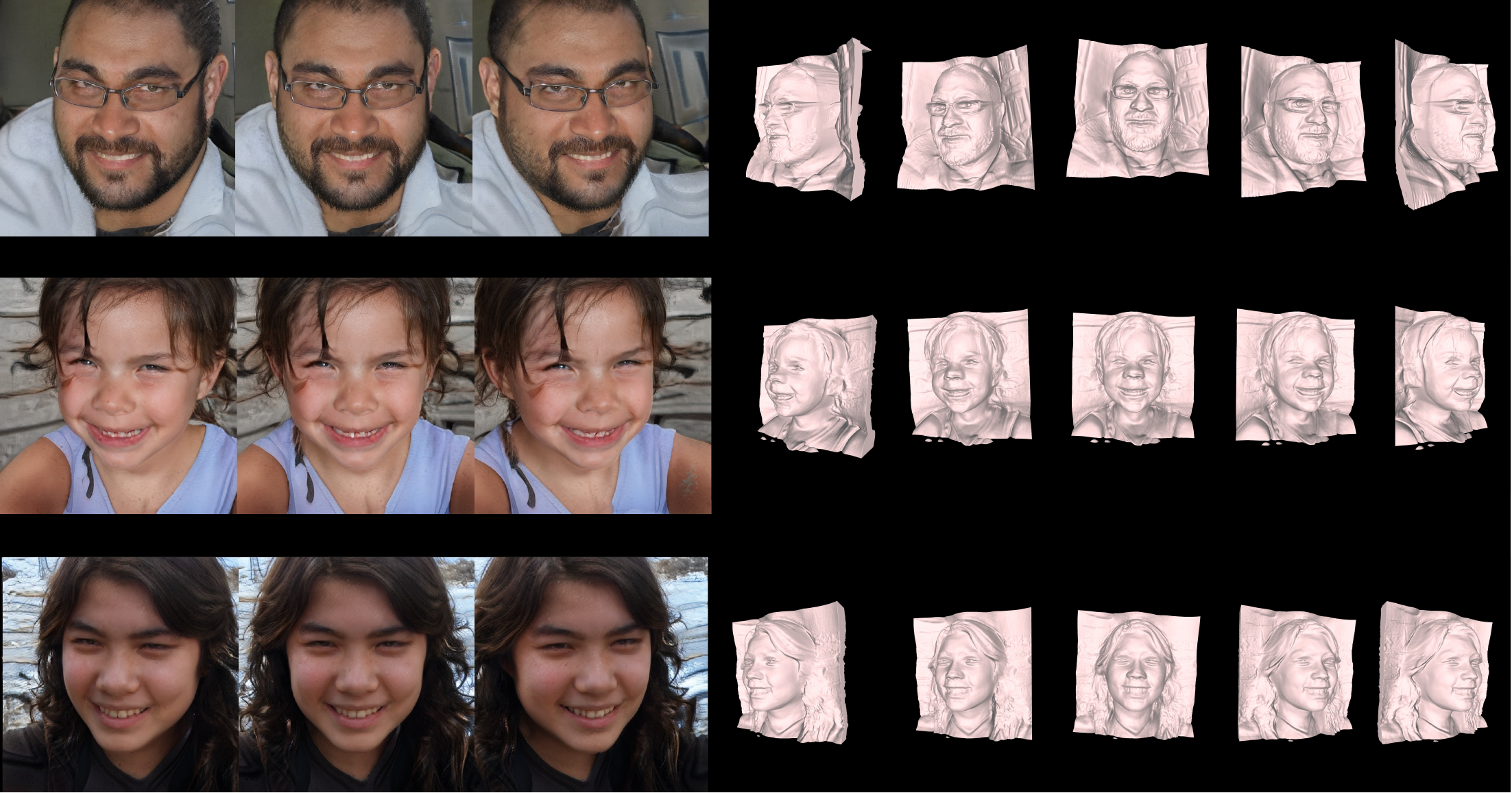}}
    \caption{
    The visualization for 2D and 3D generation result on the FFHQ dataset with EG3D+Ours.
    }
    \vspace{-.2 in}
    \label{fig:face_1}
\end{figure*}

In this section, we delve deeper into visual samples to underscore the 2D image synthesis and 3D geometric modeling capabilities of 3D-GANs, when bolstered by our method. Our results derived from ShapeNet-Cars are illustrated in Fig.~\ref{fig:car_1}. For the AFHQ dataset, the generated samples are presented in Fig.~\ref{fig:cat_2}, and the outputs corresponding to the FFHQ dataset are captured in Fig.~\ref{fig:face_1}. It is evident from these figures that 3D-GANs, trained under our proposed framework, are adept at generating high-quality results for both 2D and 3D domains. Furthermore, attached to this section are video files that offer a comparative analysis of our method against the EG3D and StyleSDF benchmarks. These videos emphasize the superior rendering of 2D images and the intricate details of mesh structures achieved by our approach. Within the videos, the methodologies of EG3D and StyleSDF are depicted, leveraging triplane and neural implicit networks respectively. It's worth noting that in the mesh representations by StyleSDF, we've addressed and amended several visual artifacts to enhance image quality. Similarly, certain discrepancies observed in the EG3D results were rectified to minimize artifacts and elevate overall visual appeal.

\section{3D Discriminative Score (3D DS) Computation}
\label{sec:dis}

Computing the 3D DS necessitates the training of a 3D classifier capable of distinguishing between various 3D objects. For this task, we source 3D objects of the target class as well as those of other classes from the ModelNet-10 dataset for training. The sampling is done in a 1:2 ratio, respectively. Once trained, this classifier aids in determining if the generated object aligns with the original class criteria.

\section{3D IoU Calculation Methodology}
\label{sec:iou}

For effective 3D IoU evaluations, we prescribe conditional inputs to the 3D-GANs. As articulated in our main text, 3D IoU computations are performed on the ShapeNet-Cars and ShapeNet-Chairs datasets, given the availability of 3D annotations therein. 
To facilitate this, an encoder is integrated to extract the image feature and subsequently project it into the latent space of the 3D-GANs. This enables the 3D-GANs to function as an image-conditional network. 
The 3D Intersection over Union (3D IoU) metric is determined by computing the intersecting and unifying volumes of the genuine and synthesized 3D bounding boxes. The resulting value is obtained by dividing the intersection volume by the union volume. This metric aptly captures the overlap across three dimensions, offering an enhanced understanding of the spatial relationship between 3D objects.

\section{User Study}
\label{sec:user}

In our paper, we highlighted the importance of the user study in evaluating the performance of our generative 3D representations, particularly the meshes. As illustrated in Fig.~\ref{fig:user}, we presented multiple side-by-side comparisons of meshes generated by our approach against those of competing methods. 
The user study solidifies our claim of superiority in the domain of 3D mesh generation. The perceptual quality, as endorsed by both experts and novices, accentuates that our approach not only aligns with technical metrics but also meets the intuitive expectations of end-users.

\begin{figure*}[t]
    \centering
    \centerline{\includegraphics[width=0.9\linewidth]{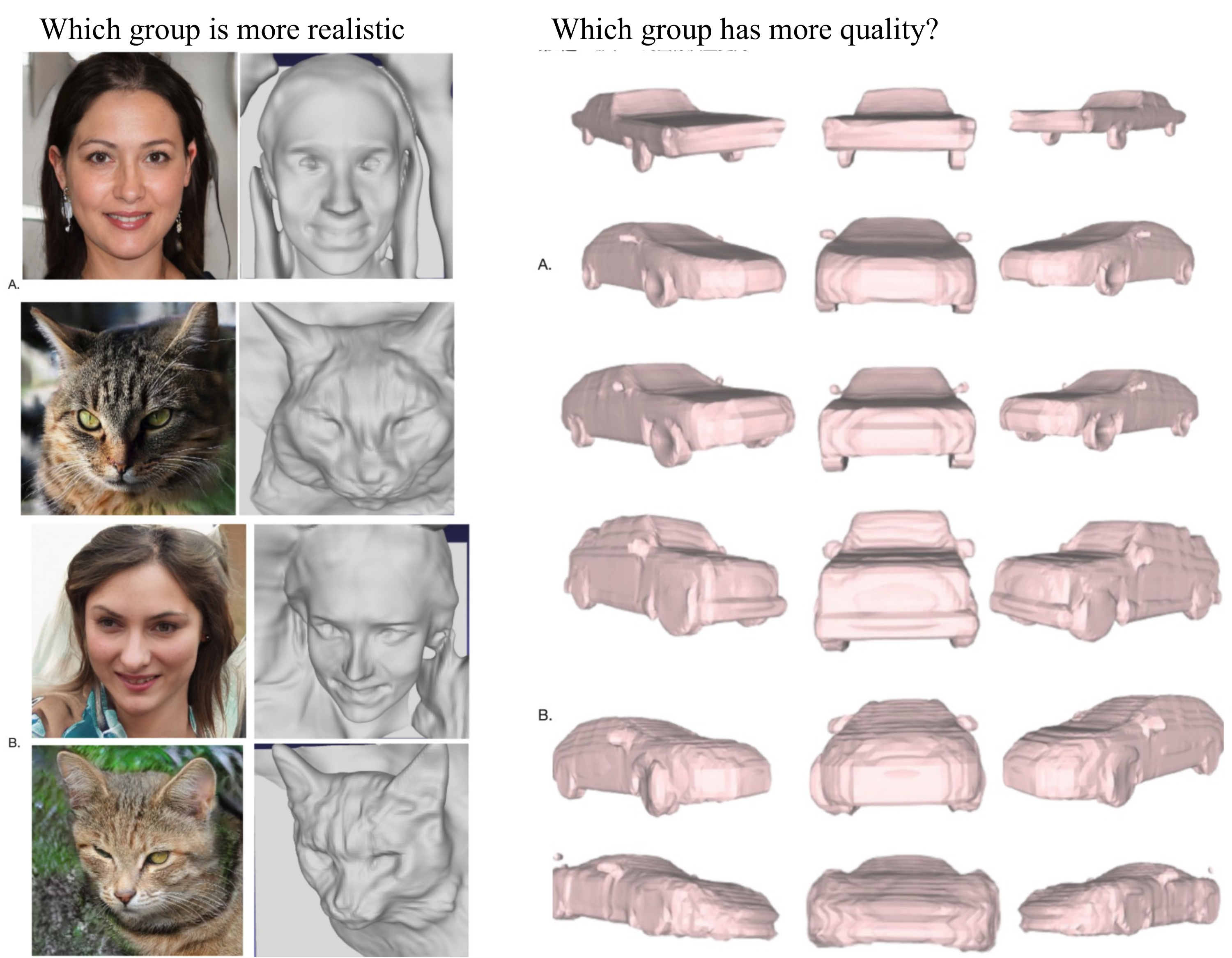}}
    \caption{
    The visualization for the user study questionnaire.
    }
    \vspace{-.2 in}
    \label{fig:user}
\end{figure*}

\end{document}